\documentclass[final]{clv2025}

\jvol{vv}
\jnum{nn}
\jyear{2026}

\dochead{Preprint} 

\usepackage{amsmath}
\usepackage{amssymb}
\usepackage{booktabs}
\usepackage{graphicx}
\usepackage{enumitem}
\usepackage{tikz}
\NewDocumentCommand{\codeword}{v}{%
\texttt{\textcolor{black}{#1}}%
}
\definecolor{cpcnavy}{HTML}{203E8C}
\definecolor{cpcteal}{HTML}{3AA594}
\definecolor{cpcorange}{HTML}{E87E37}
\definecolor{cpcred}{HTML}{732149}
\definecolor{cpcgrey}{HTML}{767676}
\definecolor{cpcline}{HTML}{D8D6D0}
\definecolor{cpcpanel}{HTML}{F7F6F3}
\definecolor{charcoalx}{HTML}{343A40}
\usetikzlibrary{positioning,arrows.meta,calc}
\renewcommand{\appendix}{%
   \setcounter{section}{0}%
   \renewcommand{\thesection}{\Alph{section}}%
   \renewcommand{\theHsection}{appx.\Alph{section}}%
   \renewcommand{\theequation}{\Alph{section}.\arabic{equation}}%
   \renewcommand{\thefigure}{\Alph{section}.\arabic{figure}}%
   \renewcommand{\thetable}{\Alph{section}.\arabic{table}}%
   \renewcommand{\theHequation}{appx.\Alph{section}.\arabic{equation}}%
   \renewcommand{\theHfigure}{appx.\Alph{section}.\arabic{figure}}%
   \renewcommand{\theHtable}{appx.\Alph{section}.\arabic{table}}%
}
\renewcommand{\appendixsection}[1]{%
   \refstepcounter{section}%
   \setcounter{table}{0}%
   \setcounter{figure}{0}%
   \setcounter{equation}{0}%
   \section*{Appendix \Alph{section}: #1}%
}

\newcommand{\Scoh}{\mathcal{S}_{\text{coh}}}

\newcommand{\Cfunc}{\mathcal{C}}

\runningtitle{A Coherentist View of Transformer Computation}
\runningauthor{Aljaafari, Freitas}

\begin{document}

\title{Align, Unify, Suppress, Route: A Coherentist View of Transformer Computation}

\author{Nura Aljaafari\thanks{Corresponding author}$^{1}$,  Andr\'{e} Freitas$^{1,2}$}

\affilblock{
    \affil{University of Manchester, United Kingdom\\\quad \email{nura.aljaafari@manchester.ac.uk}}
    \affil{Idiap Research Institute, Switzerland\\\quad \email{andre.freitas@idiap.ch}}
}

\maketitle

\begin{abstract}
Mechanistic interpretability has identified transformer circuits, but lacks a shared vocabulary for describing how their functions compose across tasks and architectures. We introduce \emph{Coherentist Probabilistic Compositionalism} (CPC), an interpretive framework that grounds transformer computation in coherentist theories of interpretation and describes it through four operator roles. \emph{Alignment} identifies candidate relations, \emph{unification} integrates supporting information, \emph{suppression} reduces incompatible alternatives, and \emph{routing} carries selected information to the output. Across 15 models from five architecture families, the suppression, unification, and routing weight-space signatures correlate with held-out activation-level role measures above random baselines. Suppression is more stable across tasks than unification. Ablating alignment heads reduces downstream suppressive activity beyond a random-head control in 10 models, but similar effects on no-conflict prompts indicate a general upstream dependency, not contradiction-specific coupling. Explicit contradictions significantly shift a layerwise coherence proxy in 14 models; after removing shared residual covariance, the gap has the predicted direction in every model. Base and instruction-tuned variants preserve induction-head score structure ($r{\geq}0.98$) without a consistent shift of operator signatures towards later layers. These results support CPC as a shared vocabulary for comparing transformer mechanisms while showing that their depth and geometric expression remain architecture-specific.\footnote{Code and data will be made publicly available upon acceptance.}
\end{abstract}

\section{Introduction}\label{sec:intro}
As transformer language models scale, their capabilities extend beyond surface pattern matching to in-context learning and other emergent behaviours \citep{olsson2022context,nanda2023,wei2022emergent}. Post-training further refines model behaviour \citep{ouyang2022instruct,bai2022constitutional}. Mechanistic interpretability seeks to explain these behaviours through the analysis of internal components and computations, often identifying circuits: sets of interacting components that contribute to a particular computation or behaviour \citep{wang2022interpretability}. Such analyses have identified induction heads, the indirect-object identification (IOI) circuit, copy-suppression heads, feedforward memories, and instruction-conditioned conflict resolution \citep{elhage2021framework,geva2021transformer,olsson2022context,wang2022interpretability,mcdougall2023copysuppression,wang2023forbiddenfacts,templeton2024scaling,aljaafari-etal-2026-emergence}. These mechanisms are well characterised individually, but how their functional roles fit within a broader framework of transformer computation is not.

Existing frameworks approach this question at different levels of abstraction. Bayesian and algorithmic analyses treat prompts as evidence for latent task inference or as inputs to implicit estimators \citep{xie2021explanation,akyurek2023icl_bayes,von2022transformers,garg2022what_can_tfs_learn}, abstracting away the components that implement the computation. Feature-level analyses decompose activations into interpretable directions \citep{elhage2022superposition,cunningham2023sparse,bricken2023monosemanticity}, primarily characterising representational content and its local causal roles, with less emphasis on how this content is combined and revised across depth. Circuit analyses identify causal components and reusable computational structure \citep{zheng2025attentionheads,he2025modcirc,aljaafari2026circuit}, but do not resolve how recurring functions can be described through graded roles that generalise across tasks and architectures. We address this gap by organising these functions into a shared vocabulary grounded in \emph{coherence refinement}: the view that a model interprets its context by linking representational fragments and progressively resolving inconsistencies among them. We test how the organisation of these functions varies across tasks, architectures, and post-training methods.

\begin{figure*}
    \centering
    \includegraphics[width=0.89\linewidth]{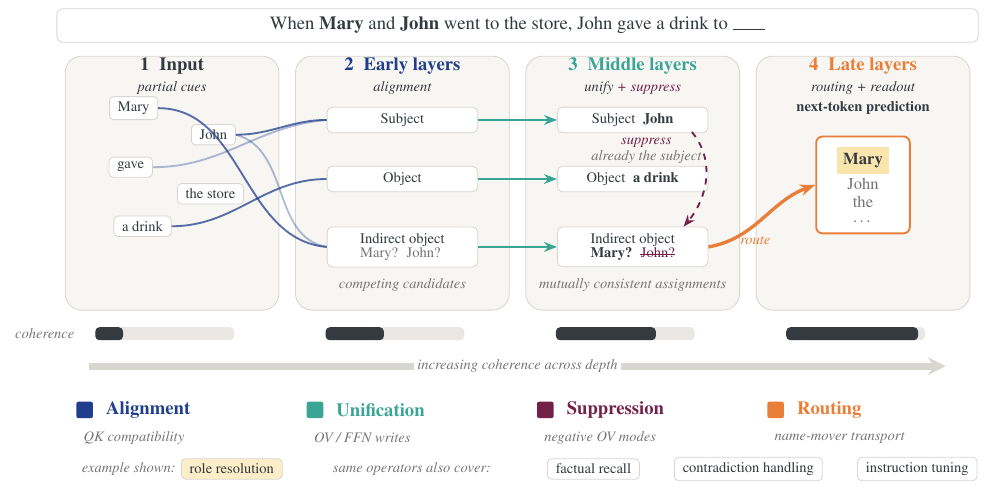}
    \caption{The CPC vocabulary on the give-frame running example. Partial cues enter, alignment proposes candidate role links, unification integrates the compatible assignments, suppression prunes the incompatible recipient, and routing conveys the consistent interpretation to the readout, with coherence increasing across depth. The legend names a characteristic mechanism for each operator, and the same operators also cover factual recall, contradiction handling, and instruction tuning. Individual mechanisms may instantiate one or more operator roles.}
    \label{fig:overview}
\end{figure*}
Since language is a structured field of inferential and referential relations, plausible continuations are constrained by discourse entities, semantic roles, and coherence relations established earlier in the context, not by surface adjacency alone \citep{halliday1976cohesion,grosz-etal-1995-centering,fillmore1982frame}. For a token-based language model, these relations must be realised through representations distributed across token positions. The representations link mentions, roles, and events, and the model must keep them compatible as the context extends (see Figure~\ref{fig:state}). Self-supervised language models encode aspects of this structure, including syntactic dependencies, semantic roles, and coreference \citep{tenney-etal-2019-bert,manning2020emergent,hu-etal-2020-systematic}. Models can draw on this structure by relating context fragments, integrating compatible information, reducing support for contradictions, and carrying the interpretation to the prediction position \citep{olsson2022context,wang2022interpretability,geva2021transformer,mcdougall2023copysuppression}.

We propose \emph{coherence refinement} as the organising principle for a vocabulary of these operations and formalise it as \emph{Coherentist Probabilistic Compositionalism} (CPC). In CPC (Figure~\ref{fig:overview}), transformer computation is modelled as the iterative refinement of representational fragments under a \emph{global coherence functional} $\Cfunc$, and components across layers may instantiate four operator roles that realise these operations: \emph{alignment}, \emph{unification}, \emph{suppression}, and \emph{routing}. Each describes what a component (e.g., an attention head or a feedforward layer) contributes to the developing interpretation: a single component may serve several roles, and one role may be spread across many components. The empirical question is how far weights and activations carry measurable signatures of these roles. We treat this as a modelling hypothesis, organising known mechanisms and generating testable predictions without assuming that transformers optimise $\Cfunc$ directly.

The empirical evaluation examines four dimensions: (i) the reproducibility and functional validity of operator signatures (P1); (ii) the dependence of suppression on upstream alignment (P2); (iii) the effect of contradiction on the coherence proxy near layers with high alignment and suppression activity (P3); and (iv) the extent to which post-training reweights operator structure or introduces new structure (P4). The paper makes three contributions:
\begin{itemize}
    \item An interpretive framework that describes transformer computation as coherence refinement carried out by four operator roles, giving mechanistic findings a shared representational language (Section~\ref{sec:cpc}).
    \item Operational definitions of attention-head operator signatures, with activation-level measures and causal tests of their proposed functional roles (Sections~\ref{sec:operators} and~\ref{sec:eval-causal}).
    \item Empirical evaluation across 15 models from five architecture families, identifying which properties of the vocabulary generalise between architectures (Section~\ref{sec:evaluation}).
\end{itemize}
\section{Coherence in Language and Cognition}\label{sec:coherence-background}
Coherence has been investigated in both philosophy and linguistics. In philosophy, theories take an epistemological view, where a belief is justified partly by the degree to which it aligns with the broader system of beliefs held by a reasoner \citep{bonjour1985structure}. In linguistics, coherence is used to characterise the relations between text units that make a discourse interpretable as a structured whole, different from an arbitrary sequence of sentences \citep{hobbs1979,jurafsky2014speech}. Despite these different objects of study, both perspectives treat coherence as a relational property: an element is evaluated by how it fits with other elements, not in isolation. 

\subsection{Coherence in Language and Discourse}\label{sec:coherence-language}
In linguistic settings, coherence is commonly studied through \emph{coherence relations}: relations that connect linguistic units such as clauses, sentences, or larger discourse spans \citep{hobbs1979}. They explain how one part of a discourse is interpreted with respect to another. For example, two spans may stand in a causal, elaborative, temporal, or contrastive relation \citep{jurafsky2014speech}. One of the most widely used theories is the Rhetorical Structure Theory, which formalises this view by representing a text through relations between discourse spans, including relations such as \textsc{Elaboration}, \textsc{Cause}, and \textsc{Contrast} \citep{mann1988rhetorical}.

\citet{jurafsky2014speech} distinguish between local and global coherence. Local coherence refers to how nearby parts of a text are connected. It has three main sources: (i) coherence relations between neighbouring text spans, (ii) continuity of discourse entities, where successive spans remain centred on the same people, objects, or events, and (iii) lexical or semantic continuity, where neighbouring spans remain within a compatible topic or semantic field. Global coherence concerns the organisation of the discourse over longer stretches of text, including how individual sentences and local relations contribute to the overall discourse structure.

Several computational approaches formalise these sources separately. Cohesion models describe surface links created through repetition, reference, and lexical relations \citep{halliday1976cohesion}. Entity-based models describe how discourse entities are introduced and maintained. For instance, Centring Theory tracks the changing focus of attention across utterances and predicts greater coherence when transitions preserve salient discourse entities \citep{grosz-etal-1995-centering}. On the other hand, Semantic approaches measure compatibility between neighbouring textual representations. More recent computational models learn such compatibility from coherent and reordered text, producing a score that distinguishes well-formed discourse from less coherent alternatives \citep{li2014model}. These approaches differ in what constitutes a discourse unit and how compatibility is measured, but all describe coherence in terms of relations among parts of the text.

An important consequence is that semantic opposition does not itself imply incoherence. A discourse may contain disagreement, contrast, correction, or concession while remaining coherent if the relation between the opposing contents is represented explicitly. For example, \textsc{Contrast} is itself a coherence relation in Rhetorical Structure Theory \citep{mann1988rhetorical}. In what follows, \emph{incompatibility} refers instead to two assignments that cannot be jointly maintained within a single interpretation.

\subsection{Coherence as Constraint Satisfaction}\label{sec:coherence-csp}
A more general formalisation treats coherence as a constraint-satisfaction problem. \citet{thagard1998coherence} define a finite set of elements $E=\{e_1,\ldots,e_n\}$ with weighted positive and negative constraints between pairs of elements. Positive constraints connect elements that support or fit with one another, while negative constraints connect elements that conflict. A solution partitions the elements into an accepted set $\mathcal{A}$ and a rejected set $\mathcal{R}$. A positive constraint between $e_i$ and $e_j$ is satisfied when both elements receive the same status, while a negative constraint is satisfied when one is accepted and the other rejected. The coherence of a partition can then be written as
\begin{align}
W(\mathcal{A},\mathcal{R})= \sum_{(e_i,e_j)\,\mathrm{satisfied}} w_{ij},
\label{eq:coherence-csp}
\end{align}
where $w_{ij}$ is the weight associated with the constraint between $e_i$ and $e_j$. The coherence problem is to find the partition that maximises $W$.

Exact maximisation of this objective is computationally difficult. A connectionist approximation represents elements as units in a network, positive constraints as excitatory connections, and negative constraints as inhibitory connections \citep{thagard1998coherence}. Mutually supporting elements reinforce one another, while conflicting elements compete. The network progressively settles into a configuration in which compatible elements tend to be active together and incompatible alternatives are separated. \citet{thagard2000coherence} applies this general formulation to different cognitive problems, including explanation, perception, analogy, and decision making. The elements and constraints differ between domains, but the underlying problem remains the same: finding a configuration that satisfies as many mutually weighted constraints as possible.

This formulation provides a useful bridge between epistemological and linguistic perspectives. In discourse, the elements may correspond to propositions, entities, events, or discourse spans. Positive constraints can represent relations that support a joint interpretation, such as coreference, causal relations, or compatible semantic roles. Negative constraints can represent mutually incompatible assignments. Coherence then depends on the configuration of these relations across the interpretation.

\section{Coherentist Probabilistic Compositionalism}\label{sec:cpc}
CPC describes transformer computation through the layerwise construction of an interpretation in the residual stream. The operator roles specify functional contributions to this process, while analyses of transformer components and circuits provide the empirical basis for identifying them.

\subsection{Coherence State and Functional}\label{sec:cpc-state}
A \emph{coherence state} $A{=}\{(m,g)\}$ is a set of representational fragments $m$ paired with groundings $g$. A fragment captures a partial hypothesis about the input, such as a token span, entity mention, syntactic role, or latent feature. A grounding maps a fragment into a relational structure $G$ representing the current interpretation. Figure~\ref{fig:state} illustrates a coherence state and its groundings. We define the global coherence functional as
\begin{equation}
\small
\begin{aligned}
\Cfunc(A){=}\;&\underbrace{\mu\,\mathrm{cov}(A)}_{\substack{\text{coverage reward}\\ \text{(unification)}}}
+ \underbrace{\sum_{(m,g)\in A}\log \Scoh(g\mid m,G)}_{\substack{\text{local fit}\\ \text{(alignment, unification)}}}\\[6pt]
&- \underbrace{\frac{\lambda}{2}\!\!\sum_{\substack{(m,g),(m',g')\in A\\ (m,g)\neq(m',g')}}\!\!\kappa(m,g;m',g')}_{\substack{\text{contradiction penalty}\\ \text{(suppression)}}} ,
\end{aligned}
\label{eq:coherence-functional}
\end{equation}
where coverage is defined over grounded positions, $\mathrm{cov}(A){=}\bigl|\bigcup_{(m,g)\in A}\operatorname{dom}(g)\bigr|$, $\Scoh(g\mid m,G){\in}(0,1]$ measures how well a grounding fits the relational structure $G$, and $\kappa{\geq}0$ is a symmetric contradiction kernel that vanishes on compatible pairs. Formal definitions of fragments, groundings, admissible pairings, and the contradiction kernel are given in Appendix~\ref{app:defs}. The functional rewards coverage ($\mu{>}0$) and local compatibility while penalising incompatible assignments ($\lambda{>}0$). The coverage term favours interpretations that incorporate more of the available input, the local-fit term favours groundings compatible with the developing relational structure, and the contradiction term penalises grounded fragments that cannot be jointly maintained. Figure~\ref{fig:configuration} illustrates the resulting view of coherence as a configuration of grounded fragments connected by relations of support and incompatibility. Appendix~\ref{app:coherence} develops the relation to the formalisation of coherence in Section~\ref{sec:coherence-background}.

\begin{figure}[h]
\centering
\resizebox{\textwidth}{!}{%
\begin{tikzpicture}[
  font=\footnotesize,
  tok/.style={draw=cpcline, fill=white, rounded corners=2pt, inner xsep=4pt, inner ysep=3pt},
  frag/.style={draw=cpcnavy, fill=cpcpanel, rounded corners=3pt, inner xsep=5pt, inner ysep=4pt},
  gnode/.style={draw=cpcgrey, fill=white, rounded corners=6pt, inner xsep=5pt, inner ysep=3pt, font=\footnotesize\scshape},
  member/.style={cpcgrey!70, line width=0.5pt},
  ground/.style={-{Stealth[length=2mm]}, cpcteal, dashed, line width=0.9pt},
  rel/.style={-{Stealth[length=2mm]}, cpcgrey, line width=0.7pt},
]
\node[tok] (t1)  at (0,0)     {Mary};
\node[tok] (t2)  at (1.15,0)  {and};
\node[tok] (t3)  at (2.3,0)   {John};
\node[tok] (t4)  at (3.45,0)  {went};
\node[tok] (t5)  at (4.45,0)  {to};
\node[tok] (t6)  at (5.4,0)   {the};
\node[tok] (t7)  at (6.5,0)   {store,};
\node[tok] (t8)  at (7.7,0)   {John};
\node[tok] (t9)  at (8.85,0)  {gave};
\node[tok] (t10) at (9.8,0)   {a};
\node[tok] (t11) at (10.8,0)  {drink};
\node[tok] (t12) at (11.85,0) {to};
\node[tok] (t13) at (12.8,0)  {\_\_\_};
\node[frag] (f1) at (0.6,1.6)  {$m_1$: entity \emph{Mary}};
\node[frag] (f2) at (4.4,2.7)  {$m_2$: entity \emph{John}};
\node[frag] (f3) at (6.9,1.6)  {$m_3$: location \emph{store}};
\node[frag] (f4) at (10.7,2.7) {$m_4$: event \emph{give}(agent, recipient, theme)};
\draw[member] (t1.north) -- (f1.south);
\draw[member] (t3.north) -- (f2.south);
\draw[member] (t8.north) -- (f2.south);
\draw[member] (t7.north) -- (f3.south);
\draw[member] (t9.north) -- (f4.south);
\draw[member] (t11.north) -- (f4.south);
\node[draw=cpcline, fill=cpcpanel, rounded corners=8pt, minimum width=5.6cm, minimum height=3.4cm, anchor=south west] (G) at (14.2,0.6) {};
\node[anchor=south west, font=\footnotesize\itshape, text=cpcgrey] at (14.4,0.75) {relational graph $G$};
\node[gnode] (nmary)  at (15.1,3.4) {Mary};
\node[gnode] (njohn)  at (16.6,3.4) {John};
\node[gnode] (nstore) at (18.9,3.4) {store};
\node[gnode] (ngive)  at (16.1,1.7) {give};
\node[gnode] (ndrink) at (18.4,1.7) {drink};
\draw[rel] (ngive) -- node[above, sloped, inner sep=1.5pt, font=\scriptsize, text=cpcgrey] {recipient} (nmary);
\draw[rel] (ngive) -- node[right, inner sep=1.5pt, font=\scriptsize, text=cpcgrey] {agent} (njohn);
\draw[rel] (ngive) -- node[above, sloped, inner sep=1.5pt, font=\scriptsize, text=cpcgrey] {at} (nstore);
\draw[rel] (ngive) -- node[above, inner sep=1.5pt, font=\scriptsize, text=cpcgrey] {theme} (ndrink);
\draw[ground] (f1.north) to[out=20,in=150] (nmary.north);
\draw[ground] (f2.north) to[out=20,in=150] (njohn.north);
\draw[ground] (f3.north) to[out=35,in=150] (nstore.north);
\draw[ground] (f4.east) to[out=10,in=180] (ngive.west);
\node[anchor=west, font=\scriptsize, text=cpcgrey] at (0,-1.0) {solid grey lines mark fragment positions $U_m$};
\node[anchor=west, font=\scriptsize, text=cpcgrey] at (6.5,-1.0) {dashed teal arrows are groundings $g\colon U_m \rightharpoonup V$};
\end{tikzpicture}}
\caption{A coherence state and its groundings. Fragments $m_1$-$m_4$ capture partial hypotheses about spans of the input, grey lines mark the positions $U_m$ they cover, and dashed teal groundings map fragment positions into the relational graph $G$.}
\label{fig:state}
\end{figure}

\begin{figure}[h]
\centering
\resizebox{0.92\textwidth}{!}{%
\begin{tikzpicture}[
  font=\footnotesize,
  fg/.style={circle, draw=cpcgrey, fill=white, inner sep=1pt, minimum size=15pt, font=\scriptsize},
  sup/.style={cpcteal},
  con/.style={cpcred, dashed},
  hdr/.style={align=center, font=\footnotesize\itshape},
  reflab/.style={font=\scriptsize\itshape, text=cpcgrey, above=1pt},
  sep/.style={-{Stealth[length=2mm]}, cpcgrey, line width=0.9pt},
]
\begin{scope}[xshift=0cm]
\path[fill=cpcpanel] plot [smooth cycle, tension=0.8] coordinates
  {(-0.45,1.9) (0.1,3.2) (1.4,3.55) (2.8,3.3) (3.95,2.5) (3.65,1.0) (2.8,-0.2) (1.1,-0.35) (-0.1,0.5)};
\node[fg, fill=charcoalx!12] (m1) at (0.55,2.75) {$m_1$};
\node[fg, fill=charcoalx!12] (m2) at (2.05,2.90) {$m_2$};
\node[fg, fill=charcoalx!8]  (m3) at (0.30,1.75) {$m_3$};
\node[fg, fill=charcoalx!15] (m4) at (1.50,1.75) {$m_4$};
\node[fg, fill=charcoalx!8]  (m5) at (3.25,2.10) {$m_5$};
\node[fg, fill=charcoalx!10] (m6) at (0.90,0.45) {$m_6$};
\node[fg, fill=charcoalx!8]  (m7) at (2.45,0.45) {$m_7$};
\draw[sup, line width=0.6pt] (m1) -- (m2);
\draw[sup, line width=0.6pt] (m3) -- (m4);
\draw[sup, line width=0.6pt] (m3) -- (m6);
\draw[con, line width=0.7pt] (m1) -- (m4);
\draw[con, line width=0.7pt] (m2) -- (m4);
\draw[con, line width=0.7pt] (m2) -- (m5);
\draw[con, line width=0.7pt] (m4) -- (m7);
\draw[con, line width=0.7pt] (m5) -- (m7);
\draw[con, line width=0.7pt] (m6) -- (m7);
\node[hdr, text=cpcred] at (1.75,4.05) {fragmented /\\ competing state};
\end{scope}
\draw[sep] (4.35,1.7) -- node[reflab] {refinement} (5.45,1.7);
\begin{scope}[xshift=5.9cm]
\path[fill=cpcpanel] plot [smooth cycle, tension=0.8] coordinates
  {(-0.45,1.9) (0.1,3.2) (1.4,3.55) (2.8,3.3) (3.95,2.5) (3.65,1.0) (2.8,-0.2) (1.1,-0.35) (-0.1,0.5)};
\node[fg, fill=charcoalx!25] (m1) at (0.55,2.75) {$m_1$};
\node[fg, fill=charcoalx!30] (m2) at (2.05,2.90) {$m_2$};
\node[fg, fill=charcoalx!22] (m3) at (0.30,1.75) {$m_3$};
\node[fg, fill=charcoalx!35] (m4) at (1.50,1.75) {$m_4$};
\node[fg, fill=charcoalx!18] (m5) at (3.25,2.10) {$m_5$};
\node[fg, fill=charcoalx!25] (m6) at (0.90,0.45) {$m_6$};
\node[fg, fill=charcoalx!18] (m7) at (2.45,0.45) {$m_7$};
\draw[sup, line width=0.9pt] (m1) -- (m2);
\draw[sup, line width=0.8pt] (m1) -- (m3);
\draw[sup, line width=1.0pt] (m3) -- (m4);
\draw[sup, line width=1.0pt] (m2) -- (m4);
\draw[sup, line width=0.9pt] (m3) -- (m6);
\draw[sup, line width=0.9pt] (m6) -- (m7);
\draw[con, line width=0.9pt] (m2) -- (m5);
\draw[con, line width=0.9pt] (m5) -- (m7);
\draw[con, line width=0.9pt] (m4) -- (m7);
\node[hdr, text=cpcorange] at (1.75,4.05) {conflict exposed /\\ compatible structure strengthens};
\end{scope}
\draw[sep] (10.25,1.7) -- node[reflab] {refinement} (11.35,1.7);
\begin{scope}[xshift=11.8cm]
\path[fill=cpcteal!10] plot [smooth cycle, tension=0.8] coordinates
  {(-0.45,1.9) (0.1,3.2) (1.4,3.55) (2.8,3.3) (3.95,2.5) (3.65,1.0) (2.8,-0.2) (1.1,-0.35) (-0.1,0.5)};
\node[fg, fill=charcoalx!32] (m1) at (0.55,2.75) {$m_1$};
\node[fg, fill=charcoalx!55] (m2) at (2.05,2.90) {$m_2$};
\node[fg, fill=charcoalx!28] (m3) at (0.30,1.75) {$m_3$};
\node[fg, fill=charcoalx!58] (m4) at (1.50,1.75) {$m_4$};
\node[fg, fill=charcoalx!22] (m5) at (3.25,2.10) {$m_5$};
\node[fg, fill=charcoalx!40] (m6) at (0.90,0.45) {$m_6$};
\node[fg, fill=charcoalx!30] (m7) at (2.45,0.45) {$m_7$};
\draw[sup, line width=1.0pt] (m1) -- (m2);
\draw[sup, line width=0.7pt] (m1) -- (m3);
\draw[sup, line width=1.3pt] (m3) -- (m4);
\draw[sup, line width=1.4pt] (m2) -- (m4);
\draw[sup, line width=0.8pt] (m2) -- (m5);
\draw[sup, line width=1.0pt] (m4) -- (m5);
\draw[sup, line width=0.9pt] (m3) -- (m6);
\draw[sup, line width=1.0pt] (m4) -- (m7);
\draw[sup, line width=1.2pt] (m6) -- (m7);
\draw[sup, line width=0.7pt] (m5) -- (m7);
\node[hdr, text=cpcteal] at (1.75,4.05) {coherent\\ configuration};
\end{scope}
\draw[cpcline, rounded corners=4pt] (-0.45,-2.35) rectangle (15.65,-1.05);
\draw[cpcline] (5.2,-1.15) -- (5.2,-2.25);
\draw[cpcline] (9.3,-1.15) -- (9.3,-2.25);
\draw[sup, line width=1.1pt] (-0.15,-1.4) -- (0.55,-1.4);
\node[font=\scriptsize, anchor=west] at (0.7,-1.4) {support (compatibility)};
\draw[con, line width=1.1pt] (-0.15,-1.95) -- (0.55,-1.95);
\node[font=\scriptsize, anchor=west] at (0.7,-1.95) {conflict (incompatibility)};
\node[font=\scriptsize, anchor=west] at (5.5,-1.4) {fragment support level};
\foreach \x/\p in {5.65/5, 6.1/20, 6.55/35, 7.0/55}{
  \draw[cpcgrey, fill=charcoalx!\p] (\x,-1.95) circle (0.13);}
\node[font=\scriptsize, anchor=west] at (7.35,-1.95) {(low $\rightarrow$ high)};
\node[font=\scriptsize] at (12.45,-1.7)
  {coverage\,$\uparrow$\;\; compatibility\,$\uparrow$\;\; conflict\,$\downarrow$\;\;$\Rightarrow$\;\;$\Cfunc(A)\uparrow$};
\end{tikzpicture}}
\caption{Coherence as a representational configuration. Circles are grounded fragments, fill level shows support, teal links mark mutual support, and red dashed links mark unresolved incompatibility, not disagreement. Refinement increases coverage and compatibility and decreases conflict, raising $\Cfunc(A)$ (Section~\ref{sec:coherence-background}; \citealp{bonjour1985structure,thagard1998coherence}). All fragments can be retained provided their relations are admissible, and support stays graded.}
\label{fig:configuration}
\end{figure}

\subsection{The Four Operators}\label{sec:cpc-operators}
The four operators describe recurring functional contributions to the refinement of a coherence state. To connect these roles to transformer computation, we use the residual-stream and attention decomposition of \citet{elhage2021framework}. Transformer blocks update an additive residual stream, while attention heads separate into query-key (QK) interactions, which determine which positions attend to one another, and value-output (OV) maps, which determine what information is written from attended positions back to the residual stream. Figure~\ref{fig:operators-arch} locates the four roles on this architecture.
\begin{figure}[t]
\centering
\resizebox{0.57\textwidth}{!}{%
\begin{tikzpicture}[
  font=\footnotesize,
  tok/.style={draw=cpcline, fill=white, rounded corners=2pt, inner xsep=5pt, inner ysep=3pt},
  stream/.style={-{Stealth[length=2mm]}, cpcgrey!60, line width=1.1pt},
  oplab/.style={font=\scriptsize, inner sep=1.5pt, fill=white},
]
\node[tok] (p1) at (-0.45,4.4) {$x_1$};
\node[tok] (p2) at (-0.45,3.4) {$x_2$};
\node[tok] (p3) at (-0.45,2.4) {$x_3$};
\node[tok] (p4) at (-0.45,1.4) {$x_4$};
\node[tok] (p5) at (-0.45,0.4) {$x_T$};
\foreach \y in {4.4,3.4,2.4,1.4,0.4}{\draw[stream] (0.15,\y) -- (9.35,\y);}
\draw[cpcline, dashed] (3.1,-0.15) -- (3.1,4.95);
\draw[cpcline, dashed] (6.2,-0.15) -- (6.2,4.95);
\node[font=\scriptsize, text=cpcgrey] at (1.6,5.2) {layer $\ell$};
\node[font=\scriptsize, text=cpcgrey] at (4.65,5.2) {$\ell{+}1$};
\node[font=\scriptsize, text=cpcgrey] at (7.75,5.2) {$\ell{+}2$};
\draw[-{Stealth[length=1.8mm]}, cpcnavy, line width=0.9pt] (1.65,1.55) to[out=75,in=-75]
  node[oplab, text=cpcnavy] {align (QK scores relations)} (1.8,4.25);
\draw[-{Stealth[length=1.8mm]}, cpcteal, line width=0.9pt] (4.65,2.25) -- (4.65,1.55);
\node[oplab, text=cpcteal] at (4.65,1.9) {unify (OV/MLP write, $+$)};
\draw[-{Stealth[length=1.8mm]}, cpcred, line width=0.9pt] (7.75,1.25) -- (7.75,0.55);
\node[oplab, text=cpcred] at (7.75,0.9) {suppress ($-$)};
\draw[-{Stealth[length=1.8mm]}, cpcorange, line width=1pt] (8.35,4.25) to[out=-55,in=55]
  node[oplab, text=cpcorange, pos=0.45] {route (mover head)} (9.15,0.55);
\node[draw=cpcline, fill=white, rounded corners=2pt, inner xsep=4pt, inner ysep=2.5pt, font=\scriptsize] at (10.05,0.4) {readout};
\end{tikzpicture}}
\caption{The CPC operators on the transformer architecture, with tokens on the left and depth running left to right. Query-key structure aligns positions, OV and feedforward writes unify content, subtractive writes suppress incompatible content, and a mover head routes the result to the readout.}
\label{fig:operators-arch}
\end{figure}

\paragraph{Alignment}
It identifies candidate relations between positions or representational fragments. In transformers, QK interactions provide input-dependent compatibility scores that determine which positions exchange information.
\paragraph{Unification}
It incorporates compatible content into the developing interpretation. Once a fragment is aligned with a candidate role or relation, additive component writes can increase support for the corresponding interpretation in the residual stream without removing competing alternatives.
\paragraph{Suppression}
It reduces support for continuations, features, or fragments that are incompatible with the developing interpretation. In transformers, this role can be implemented through subtractive component writes along particular readout directions.

\paragraph{Routing}
It carries selected content to positions where it can affect the readout. It changes where information is available without necessarily introducing new representational content.

These roles correspond to different changes in the coherence state and functional. Alignment proposes candidate groundings. Unification incorporates compatible grounded fragments, increasing coverage and local fit. Suppression reduces support for alternatives that contribute to the contradiction penalty. Routing does not directly change $\Cfunc$ and instead controls where the resulting content is available to affect the model's prediction.

\subsection{Layerwise Coherence Refinement}\label{sec:cpc-ascent}
CPC interprets transformer depth as a sequence of approximate coherence-refinement steps in which the operators act on the current representational state. Figure~\ref{fig:overview} illustrates this process on the running example. A stronger conjecture is that the latent coherence functional tends to improve across depth on typical inputs:
\begin{equation}
\Cfunc\!\left(A^{(\ell+1)}\right)
\gtrsim
\Cfunc\!\left(A^{(\ell)}\right).
\label{eq:ascent}
\end{equation}
We treat Equation~\ref{eq:ascent} as a falsifiable conjecture, not as an assumption of monotonic improvement at every layer. Operators may overlap, repeat, or interact across layers, and CPC does not impose a fixed sequence in which alignment must precede unification, suppression, or routing.

\subsection{Transformer Mechanisms as CPC Operators}\label{sec:operators}
Known circuits and transformer mechanisms can be described through the four operator roles, although the correspondence is not one-to-one. Circuit findings are measured through attention patterns, activations, ablations, and logit contributions, while the operator labels describe functional contributions at a higher level of abstraction. A single mechanism may combine several roles, and one role may be implemented by several components.
\begin{table}[t]
\centering
\footnotesize
\setlength{\tabcolsep}{3pt}
\renewcommand{\arraystretch}{1.12}
\begin{tabular}{@{}lp{0.43\linewidth}p{0.42\linewidth}@{}}
\toprule
\textbf{Operator} &
\textbf{Mechanistic signature} &
\textbf{Representative circuits} \\
\midrule
Alignment & QK compatibility structure & Previous-token, duplicate-token heads \\
Unification & Positive OV or feedforward writes & Induction-head OV; key-value memories \\
Suppression & Negative OV modes; subtractive writes & S-inhibition, copy-suppression heads \\
Routing & Content-preserving OV structure; positional transport (activations) & Name-mover heads \\
\bottomrule
\end{tabular}
\caption{CPC operators, their candidate mechanistic signatures, and representative circuits associated with each role; sources are cited in the text.}
\label{tab:cpc-operators}
\end{table}

Table~\ref{tab:cpc-operators} summarises the proposed correspondence. Induction combines alignment and unification: a previous-token head supplies shifted positional context, and the induction head copies a compatible continuation through its OV map \citep{olsson2022context}. The IOI circuit combines all four roles: duplicate-token heads align repeated names, S-inhibition heads suppress the duplicated-subject pathway, and name-mover heads route the correct name to the prediction position while their OV writes increase support for that name \citep{wang2022interpretability}. Copy-suppression heads provide a direct example of subtractive writes \citep{mcdougall2023copysuppression}, while feedforward key-value memories provide an example of content integration \citep{geva2021transformer}. Instruction-conditioned suppression connects the vocabulary to post-training \citep{wang2023forbiddenfacts}. Sparse autoencoder features offer a candidate measurement basis for CPC fragments \citep{cunningham2023sparse,templeton2024scaling}, although the correspondence remains partial.

\paragraph{Per-head operator scores}
For each head $h$, we derive weight-space quantities corresponding to the proposed mechanistic roles. Let $W_{OV}=W_O W_V$ denote the effective OV map, with singular value decomposition
$W_{OV}=\sum_i \sigma_i u_i v_i^\top$.
For residual state $r$ and token readout direction $e_t$, its contribution to the token logit is
\begin{equation}
e_t^\top W_{OV}\,r
=
\sum_i
\sigma_i
\langle e_t,u_i\rangle
\langle v_i,r\rangle .
\label{eq:svd-logit}
\end{equation}
A mode contributes subtractively to token $t$ when the corresponding term in Equation~\ref{eq:svd-logit} is negative. This motivates separating additive and subtractive OV structure at the weight level, while task-specific effects still depend on the residual state $r$ and readout direction $e_t$. We compute
\begin{equation}
\begin{aligned}
s_{\mathrm{align}}(h)
&=
\bigl(\sigma^{QK}_{h,1}\bigr)^2 \big/ {\textstyle\sum_i} \bigl(\sigma^{QK}_{h,i}\bigr)^2, \\
s_{\mathrm{sup}}(h)
&=
{\textstyle\sum_i} \sigma^{OV}_{h,i}\,\max\!\left(0,-\langle u^{OV}_{h,i},v^{OV}_{h,i}\rangle\right), \\
s_{\mathrm{unify}}(h)
&=
{\textstyle\sum_i} \sigma^{OV}_{h,i}\,\max\!\left(0,\langle u^{OV}_{h,i},v^{OV}_{h,i}\rangle\right), \\
s_{\mathrm{trace}}(h)
&=
\tfrac{1}{d}\,\mathrm{tr}(W_{OV,h}),
\end{aligned}
\label{eq:head-scores}
\end{equation}
where $d{=}d_{\mathrm{model}}$, $\sigma^{QK}_{h,i}$ are the singular values of the QK map, and $\sigma^{OV}_{h,i}$, $u^{OV}_{h,i}$, and $v^{OV}_{h,i}$ are the singular values and vectors of the OV map. The alignment score measures QK spectral concentration, while the suppression and unification scores quantify subtractive and additive OV modes. Since
\[
\mathrm{tr}(W_{OV,h})=\sum_i\sigma^{OV}_{h,i}\langle u^{OV}_{h,i},v^{OV}_{h,i}\rangle,
\]
the trace descriptor satisfies $s_{\mathrm{trace}}=(s_{\mathrm{unify}}-s_{\mathrm{sup}})/d$ exactly and is not treated as an independent signature coordinate.

Routing is instead represented by the embedding-space copy score of \citet{elhage2021framework}. For a seeded token sample with effective embeddings collected in $E$ (token embedding plus the first-layer MLP output; \citealp{mcdougall2023copysuppression}) and unembedding directions in $U$, the map $M_h=E\,W_{OV,h}\,U$ measures how strongly head $h$ maps token representations towards their corresponding readout directions. We define
\begin{equation}
s_{\mathrm{copy}}(h)=\frac{\sum_i \max\{0,\lambda_i(M_h)\}}{\sum_i |\lambda_i(M_h)|}\in [0,1],
\label{eq:copy-score}
\end{equation}
where $\lambda_i$ are the eigenvalues of $M_h$. A head that preserves token identity towards the readout scores near one, while a head dominated by inverted mappings scores near zero. The copy score depends on the embedding geometry and is not an algebraic function of the other scores. The resulting signature is $(s_{\mathrm{align}},s_{\mathrm{sup}},s_{\mathrm{unify}},s_{\mathrm{copy}})$.

These signatures are defined for attention heads. Feedforward contributions to unification (Table~\ref{tab:cpc-operators}) enter CPC through the functional mapping and are not assigned separate weight-space scores in the present evaluation. Position-dependent routing is also measured separately at the activation level. The signatures nominate candidate operator roles whose functional validity is tested in Section~\ref{sec:evaluation} using held-out activation measures and causal ablations. CPC does not assume that every component belongs to one operator class, that every task requires all four roles, or that transformers explicitly optimise Equation~\ref{eq:coherence-functional}.

\subsection{Activation-Level Coherence Proxy}\label{sec:cpc-proxy}
The functional $\Cfunc$ is latent, so the empirical analysis uses an activation-level proxy for one possible geometric consequence of coherence:
\begin{equation}
\hat{\Cfunc}_{\mathrm{link}}^{(\ell)} =
\frac{1}{|P|}\sum_{(i,j)\in P}
\mathrm{sim}\!\left(r_i^{(\ell)},\, r_j^{(\ell)}\right),
\label{eq:coherence-proxy}
\end{equation}
where $r_i^{(\ell)}$ is the residual-stream vector at position $i$ and layer $\ell$, $P$ is a set of position pairs, and $\mathrm{sim}$ is a bounded similarity function. Higher similarity is treated as a candidate signature of representational coherence across the prompt, and an explicit contradiction is predicted to reduce it.

The proxy captures only one geometric consequence that may accompany changes in $\Cfunc$ and does not uniquely isolate relational coherence. We compare it against matched non-contradictory and shuffled position-pair controls, and evaluate a whitened variant that reduces the contribution of global residual-stream covariance.
The contradiction experiment compares matched consistent and contradictory prompts across depth (P3, Section~\ref{sec:eval-contradiction}), while the ascent analysis tests the relation between the proxy and layer depth on mixed prompts (Section~\ref{sec:cpc-ascent}). The experimental choices of $P$ and $\mathrm{sim}$ are specified in Section~\ref{sec:eval-setup}.

\subsection{Post-Training as Operator Reweighting}\label{sec:rlhf}
Post-training methods, including supervised instruction tuning and reinforcement learning from human feedback (RLHF), modify model behaviour after pretraining. The mechanistic question is how much these changes alter the organisation of computations already present in the base model. RLHF provides one formal instance: it is commonly expressed as reward optimisation under a KL constraint to a reference policy \citep{ouyang2022instruct,bai2022constitutional}. In CPC terms, the preference objective changes the relative importance of behaviours expressed through the operator roles, while the KL constraint limits divergence from the pretrained computation. This motivates testing if post-training primarily reweights existing operator structure.

\section{Methodology}\label{sec:eval-setup}
The experiments test four predictions: operator signatures in weight space (P1), suppression-alignment coupling (P2), coherence under contradiction (P3), and operator reweighting under post-training (P4). Layerwise composition and cross-task stability serve as secondary analyses.

\paragraph{Models}
We evaluate 15 models from five families: GPT-2 \citep[Small, Medium, Large;][]{radford2019language}, Pythia \citep[410M, 1.4B, 2.8B;][]{biderman2023pythia}, Qwen~2.5 \citep[1.5B, 3B;][]{yang2024qwen25}, Gemma~2 \citep[2B;][]{gemmateam2024gemma2}, and LLaMA~3.2 \citep[1B, 3B;][]{grattafiori2024llama3}, plus the available instruction-tuned variants (Gemma, LLaMA-1B, LLaMA-3B, Qwen-1.5B). The P1 weight-space analyses operate on model weights without inference. Experiments using prompts cover all models, except P4, which compares the four base/instruction-tuned pairs.

\paragraph{Tasks and data}
We use indirect object identification \citep[IOI;][]{wang2022interpretability}, greater-than comparison \citep{hanna2023greaterthan}, and factual recall \citep{meng2022rome}, formally defined in Appendix~\ref{app:tasks}. For each task, examples are filtered to those the model predicts correctly, with a target of 500 examples per model-task pair. Greater-than requires the two-digit year continuation to be a single token, which the LLaMA, Qwen, and Gemma tokenisers do not provide; the affected models are excluded from this task only. Each task set is split 50/50 into discovery and held-out subsets using a seeded hash fixed at generation time. Head selection and ranking are performed on the discovery split before held-out evaluation. Contradiction experiments use 500 matched consistent/contradictory prompt pairs per run, sampled with a fixed seed from a frozen pool of 3{,}854 unique pairs (generation templates and pool composition in Appendix~\ref{app:contradiction-templates}).

\paragraph{Operational measures}
CPC is evaluated through weight-space and activation-level measures. In weight space, each attention head is represented by the four signature coordinates defined in Section~\ref{sec:operators}, $(s_{\mathrm{align}}, s_{\mathrm{sup}}, s_{\mathrm{unify}}, s_{\mathrm{copy}})$. At the activation level, unification and suppression are measured through positive and negative direct logit attribution \citep[DLA;][]{nostalgebraist2020} over the model's top-10 predicted tokens. Routing is measured through an attention-weighted copy score, and alignment through the maximum previous-token or duplicate-token attention on repeated random sequences \citep{olsson2022context,wang2022interpretability}. Layerwise coherence is measured using $\hat{\Cfunc}_{\mathrm{link}}^{(\ell)}$ (Equation~\ref{eq:coherence-proxy}), instantiated as the mean cosine similarity over all position pairs in the residual stream, excluding the beginning-of-sequence token. Task-based activation measures are averaged over correctly predicted examples. For comparison across architectures of different depths, layerwise measures are aggregated into early, middle, and late thirds of the layer stack (Appendix~\ref{app:params}).

\paragraph{P1: operator signatures}
We first characterise the geometry of the four-dimensional weight-space signature. Within each model, heads from all layers are pooled and each coordinate is standardised. To identify recurring signature profiles, we apply $k$-means clustering \citep{macqueen1967some} with 50 restarts for $k \in \{2,\ldots,7\}$. To characterise continuous axes of variation, we apply principal component analysis \citep[PCA;][]{pearson1901lines} to the same head-by-coordinate matrix. The algebraically derived trace descriptor is excluded from both analyses. Pooling is used because P1 concerns model-wide operator structure, and the influence of depth is quantified separately through the depth-separation ratio and depth residualisation. Clustering quality is measured by the silhouette coefficient \citep{kaufman1990finding} and calibrated against two nulls in the same four-coordinate space: a column-shuffle null, which preserves each coordinate's marginal distribution while removing cross-score dependence, and an isotropic standard Gaussian null with the same number of heads and dimensions.

We test functional validity by correlating the suppression, unification, and routing weight-space scores (Spearman) with their corresponding held-out activation-level measures. The correlations are compared against the 95\% quantile of 100 random weight projections, weight-magnitude ranking, and, for suppression and unification, scores computed after shuffling task relations. Alignment is evaluated separately through its activation-level detector. We also correlate the weight-space scores with held-out single-head ablation effects, defined as the absolute change in the correct-token logit after zero-ablating a head. This analysis uses the union of the top-10 heads under each operator ranking, so a positive correlation indicates that higher-scoring heads have larger causal effects irrespective of direction.

For direct causal validation, we ablate the top five heads per signature dimension on IOI, using the same selection rule across models, and measure changes in the correct-token (IO) logit, wrong-token (subject) logit, and logit margin. We compare these effects against layer-matched random-head ablations (20 control samples per intervention) and a magnitude-matched control. As a task-conditioned check, suppression and unification heads are additionally ranked by the targeted DLA margin on the discovery split and ablated on the held-out split.

\paragraph{P2: suppression-alignment coupling}
Ablating upstream alignment heads should reduce the summed suppressive contribution $s^{\mathrm{act}}_{\mathrm{sup}}(h,x,t)=\max\bigl(0,-e_t^\top o^{(h)}_{i^\ast}(x)\bigr)$ of downstream suppressor heads $h$, where $o^{(h)}_{i^\ast}(x)$ is the head output at the readout position and $e_t$ the unembedding direction of the incompatible candidate $t$. The candidate is the top-1 continuation of the matched consistent prompt, rendered incompatible by the inserted contradiction. On contradiction prompts, suppression is compared before and after alignment-head ablation against two controls: layer-matched random-head ablation and the same intervention on matched no-conflict prompts. Significance is assessed by paired permutation tests (10{,}000 permutations), Holm-corrected \citep{holm1979simple} across the 30 model-by-contrast comparisons, with percentile bootstrap confidence intervals \citep{efron1979bootstrap} based on 10{,}000 resamples. GPT-2~Small uses the validated IOI circuit, with duplicate- and previous-token heads as alignment heads, and S-inhibition and negative name-mover heads as suppressors. Elsewhere, alignment heads are the five heads scoring highest on the activation-level alignment measure, which uses repeated random sequences and no task data, and suppressor heads are the six heads with the most-negative mean DLA on $t$, estimated on the discovery split of the contradiction prompts.

\paragraph{P3: coherence under contradiction}
The coherence gap is the consistent-minus-contradictory difference in $\hat{\Cfunc}_{\mathrm{link}}^{(\ell)}$ on matched prompt pairs. It is tested per layer third using paired $t$-tests with Holm correction, and checked for co-location with the layer third in which the model's alignment or suppression activity peaks. Three controls qualify the raw gap. Lexically matched non-contradictory pairs isolate the contradiction substitution, while a shuffled position-pair control, which pairs positions across prompts, separates within-prompt relational structure from a global state shift. A per-layer whitened proxy applies shrinkage-regularised ZCA whitening \citep{kessy2018optimal} to residual vectors before cosine similarity, with covariance shrunk towards the average-variance identity ($\alpha{=}0.1$), and is used to test the directional prediction. The layerwise logit-lens rank trajectory \citep{nostalgebraist2020} of the coherent continuation provides a scale-free convergent measure, and a naturalistic Wikipedia prompt set probes out-of-template replication.

\paragraph{P4: post-training reweighting}
The base/instruction-tuned pairs are compared through per-head operator signatures in the early and late halves of the network. We additionally measure the Pearson correlation of induction scores for corresponding heads between variants. Preservation is assessed using Fisher's $z$-transformation \citep{fisher1915frequency} with the one-sided hypotheses $H_0{:}\ \rho {\leq} 0.9$ and $H_1{:}\ \rho {>} 0.9$.

\paragraph{Secondary analyses}
Layerwise composition averages the per-head activation-level measures by layer third on correctly predicted prompts. A targeted variant contrasts compatible and incompatible tokens through the per-head DLA margin (indirect object vs.\ subject in IOI, answer vs.\ distractor in factual recall). Cross-task stability correlates per-head DLA-based unification and suppression scores between task pairs, subject to the task exclusions above.

\paragraph{Decision rules and statistical testing}
The following decision rules were fixed before the runs. Outcomes satisfying none of the stated \textsc{support}, \textsc{partial}, or \textsc{mixed} criteria are labelled \textsc{against}. \emph{P1 (signatures):} \textsc{support} requires the mean of the corresponding functional correlations to exceed all applicable baselines and the ablation-effect correlation to be positive; \textsc{partial} requires at least one applicable baseline to be exceeded. The optimal $k$ is the silhouette peak over $k \in \{2,\ldots,7\}$. \emph{P2 (coupling):} \textsc{support} requires alignment-head ablation to reduce summed suppressive contribution more than random-head ablation (Holm-corrected $\alpha{=}0.05$) and the reduction to be weaker on matched no-conflict prompts; \textsc{partial} requires only the first condition. \emph{P3 (contradiction sensitivity):} \textsc{support} requires a significant raw coherence gap in at least one layer third (Holm-corrected), with the largest absolute gap co-locating with the model's peak alignment or suppression third (Section~\ref{sec:eval-layered}); \textsc{mixed} denotes a significant but non-co-located gap. The directional prediction is scored separately on the whitened proxy, where \textsc{support} requires a significant positive consistent-minus-contradictory gap. \emph{P4 (base vs.\ tuned):} \textsc{support} requires the late suppression-and-routing shift to exceed the early alignment shift and induction preservation to pass the one-sided test with threshold $r{=}0.9$; \textsc{partial} requires one condition. \emph{Layerwise composition:} \textsc{support} requires unification to peak in the middle or late third, routing in the late third, and suppression to increase from early to middle; \textsc{partial} requires two of three conditions. \emph{Cross-task stability:} \textsc{support} requires mean cross-task $r{>}0.5$ for both suppression and unification; \textsc{partial} requires the threshold for one operator.

\paragraph{Software and implementation}
Experiments were run on NVIDIA RTX A6000 GPUs with Python~3.11.15. Core dependencies are NumPy~2.4.6 \citep{harris2020array}, PyTorch~2.7.1 with CUDA~12.6 \citep{paszke2019pytorch}, TransformerLens~3.5.1 \citep{nanda2022transformerlens}, Transformers~5.14.1 \citep{wolf2020transformers}, scikit-learn~1.9.0 \citep{scikit-learn}, SciPy~1.17.1 \citep{virtanen2020scipy}, and h5py~3.16.0.

\section{Results and Discussion}\label{sec:evaluation}
\subsection{Operator Signatures (P1)}\label{sec:eval-clustering}
CPC predicts reproducible functional information in the operator scores $(s_{\mathrm{align}}, s_{\mathrm{sup}}, s_{\mathrm{unify}}, s_{\mathrm{copy}})$ (Section~\ref{sec:operators}), without assuming that heads form four disjoint operator classes.

\begin{table}[t]
\centering
\footnotesize
\begin{tabular}{@{}lccccccc@{}}
\toprule
\textbf{Model} & \textbf{Sil.} & \textbf{Best $k$} & \textbf{$p_{\mathrm{shuf}}$} & \textbf{W$\to$Act $r$} & \textbf{Magn.\ $r$} & \textbf{Bl.} & \textbf{Abl.} \\
\midrule
GPT-2 Small & 0.42 & 3 & $0.15$ & 0.54 & 0.22 & 3/3 & -- \\
GPT-2 Medium & 0.46 & 3 & ${<}0.01$ & 0.54 & 0.11 & 3/3 & \checkmark \\
GPT-2 Large & 0.42 & 3 & ${<}0.01$ & 0.58 & 0.15 & 3/3 & \checkmark \\
Pythia-410M & 0.36 & 5 & $0.82$ & 0.30 & 0.07 & 3/3 & \checkmark \\
Pythia-1.4B & 0.42 & 3 & ${<}0.01$ & 0.34 & 0.05 & 3/3 & -- \\
Pythia-2.8B & 0.36 & 6 & ${<}0.01$ & 0.33 & 0.10 & 3/3 & -- \\
Qwen-1.5B & 0.48 & 2 & $0.12$ & 0.21 & 0.25 & 2/3 & -- \\
Qwen-3B & 0.50 & 2 & ${<}0.01$ & 0.23 & 0.37 & 2/3 & -- \\
Gemma-2B & 0.49 & 5 & ${<}0.01$ & 0.42 & 0.07 & 3/3 & -- \\
LLaMA-1B & 0.50 & 3 & ${<}0.01$ & 0.32 & 0.23 & 3/3 & -- \\
LLaMA-3B & 0.49 & 4 & ${<}0.01$ & 0.29 & 0.35 & 2/3 & -- \\
Qwen-1.5B-It & 0.48 & 2 & $0.12$ & 0.21 & 0.24 & 2/3 & -- \\
Gemma-2B-It & 0.49 & 5 & ${<}0.01$ & 0.45 & 0.07 & 3/3 & -- \\
LLaMA-1B-It & 0.50 & 4 & ${<}0.01$ & 0.26 & 0.27 & 2/3 & -- \\
LLaMA-3B-It & 0.50 & 4 & ${<}0.01$ & 0.32 & 0.14 & 3/3 & -- \\
\bottomrule
\end{tabular}
\caption{Operator-signature structure and functional validation (means over 5 seeds), computed on the four independent signature coordinates (alignment, suppression, unification, routing-copy). Sil.: silhouette at $k{=}4$; Best $k$: silhouette-optimal cluster count; $p_{\mathrm{shuf}}$: maximum column-shuffle null $p$-value across seeds; W$\to$Act $r$: held-out Spearman correlation between weight scores and activation-level role measures, against the weight-magnitude baseline (Magn.); Bl.: baselines beaten (random-projection, shuffled-relation, weight-magnitude); Abl.: positive rank correlation with held-out single-head ablation effects. The pre-specified P1 verdict is \textsc{support} with 3/3 baselines and a positive ablation correlation (three models), \textsc{partial} otherwise (twelve).}
\label{tab:exp1}
\end{table}

\paragraph{Signature scores correlate with held-out functional roles}
On held-out inputs, the suppression, unification, and routing scores correlate with the matching activation-level role measures at mean Spearman $r$ between $0.21$ and $0.58$ per model, positive throughout (Table~\ref{tab:exp1}). The copy score is the strongest single predictor ($r{=}0.59$ against the activation-level copy measure, versus $0.40$ for the derived trace score). The correlation exceeds the random-projection and shuffled-relation baselines in all models, but the weight-magnitude baseline in only 10: in the Qwen family and two LLaMA variants, generic OV magnitude tracks activation-level roles as well as the structured scores do, \emph{suggesting that these families concentrate functional roles in their largest-norm heads}. However, the alignment weight score does not track the attention-based alignment measure ($r{\approx}{-}0.04$); evidence for the alignment role accordingly depends on the activation-level detection score. Predicting held-out single-head ablation effects is harder, with a positive rank correlation in 3 models. The pre-specified rule gives 3 \textsc{support} and 12 \textsc{partial} verdicts, with no model \textsc{against}: \emph{the signatures carry functional role information beyond chance in every model and beyond naive magnitude in two-thirds}, while reliable prediction of causal effect sizes remains the open gap.

\paragraph{Structured signatures appear across all families, but the optimal cluster count varies}
Silhouette coefficients at $k{=}4$ range from $0.36$ to $0.50$ (Table~\ref{tab:exp1}). A two-cluster separation between QK- and OV-specialised heads provides a natural baseline, with higher $k$ indicating finer role-dominant profiles. The preferred cluster count remains architecture-dependent: $k{=}2$ across the Qwen family, $k{=}3$ in the GPT-2 family, Pythia-1.4B, and LLaMA-1B, $k{=}4$ in the remaining LLaMAs, and $k{=}5$ or $k{=}6$ in Gemma, Pythia-410M, and Pythia-2.8B. No single cluster count describes all families, and no family cleanly reproduces a four-profile organisation. \emph{This supports the framework's position that operator roles are graded properties of heads, not disjoint classes} (Section~\ref{sec:operators}): most heads mix additive and subtractive OV modes, and the preferred count tracks how sharply a family separates its few specialised heads from this mixed background.

\paragraph{Cross-score structure exceeds marginal-preserving nulls in most models}
Two nulls calibrate the silhouettes (Table~\ref{tab:exp1}; Section~\ref{sec:eval-setup}). Every model exceeds the Gaussian null ($p{<}0.01$ against 200 draws; mean $\Delta{=}+0.28$), and 11 exceed the column-shuffle null at $p{<}0.01$ in every seed (mean $\Delta{=}+0.09$, exceptions GPT-2~Small, Pythia-410M, and Qwen-1.5B variants). The joint structure is carried largely by the copy score's coupling to the OV mode balance ($r(s_{\mathrm{sup}}, s_{\mathrm{copy}}){=}{-}0.61$, $r(s_{\mathrm{unify}}, s_{\mathrm{copy}}){=}{+}0.44$): heads dominated by subtractive OV modes anti-copy token identity. Clusters track depth only partially: the depth-separation ratio (between-band share of signature variance) is $0.02$--$0.24$ (mean $0.08$), and regressing out layer position leaves substantial clusterability (best-$k$ silhouette $0.35$--$0.56$), changing the preferred count in 5 models. Base and instruction-tuned variants produce nearly identical silhouettes (difference ${\leq} 0.001$; best $k$ unchanged except LLaMA-1B, $3{\to}4$), suggesting limited weight-space reorganisation.

\paragraph{The signatures separate QK specialisation from OV mode balance}
PCA on the four standardised coordinates gives components of $49\%$, $27\%$, $18\%$, and $7\%$: no single axis dominates. Alignment is nearly uncorrelated with every other score ($|r| \leq 0.07$) and suppression and unification are moderately anti-correlated ($r{=}{-}0.30$). QK specialisation hence varies independently of an OV subspace in which mode balance and copying are coupled. Functionally, the suppression, unification, and copy scores predict their matching activation-level measures at $r{=}{+}0.21$, $+0.28$, and $+0.59$. The geometry separates the signature space into two nearly independent families, a QK axis and an OV subspace, and explains why the alignment role needs its own activation-level measure: \emph{the additive or subtractive character of an OV map is fixed by the weights, while QK structure acts only in combination with the input; the present weight-space alignment score does not recover the activation-level attention pattern}.

\subsection{Layerwise Operator Composition}\label{sec:eval-layered}
Following Section~\ref{sec:cpc-ascent}, which interprets depth as coherence refinement, CPC expects a depth-stratified distribution of operator roles: alignment early, unification and suppression mid-network, routing late. The discriminative prediction is that suppression should not follow the generic late-peaking pattern.

Suppression peaks in the middle third for Pythia-1.4B and Pythia-2.8B, and in the early third for Pythia-410M, GPT-2~Large, and both Gemma variants; the decomposition distinguishes operator timing beyond the generic late-layer bias. The full decision rule (unification peaking mid-or-late, routing late, suppression increasing early-to-mid) holds in 8 models; the remaining 7 satisfy two of three conditions. Gemma remains the most divergent family: suppression and alignment peak in the early third while routing peaks late, suggesting early suppressive processing followed by conventional late readout. Operators distribute non-uniformly across depth and suppression is separable from the late-concentration baseline, but the three-stage partition (align/suppress/route) does not hold cleanly: unification and routing co-occur in the later part, in line with content becoming readout-aligned only late in the model, and suppression timing is architecture-dependent. \emph{The three-stage organisation of depth holds as a tendency, not a strict sequence.} The targeted DLA variant (Section~\ref{sec:eval-setup}) concentrates the correct-token margin in the late third of both tasks (GPT-2~Small IOI: early $0.00$, late $+0.08$ per head), with factual margins an order of magnitude smaller and no support for an early factual margin.

\subsection{Causal Ablation of Operator-Ranked Heads}\label{sec:eval-causal}
\begin{figure}[t]
  \centering
  \includegraphics[width=\linewidth]{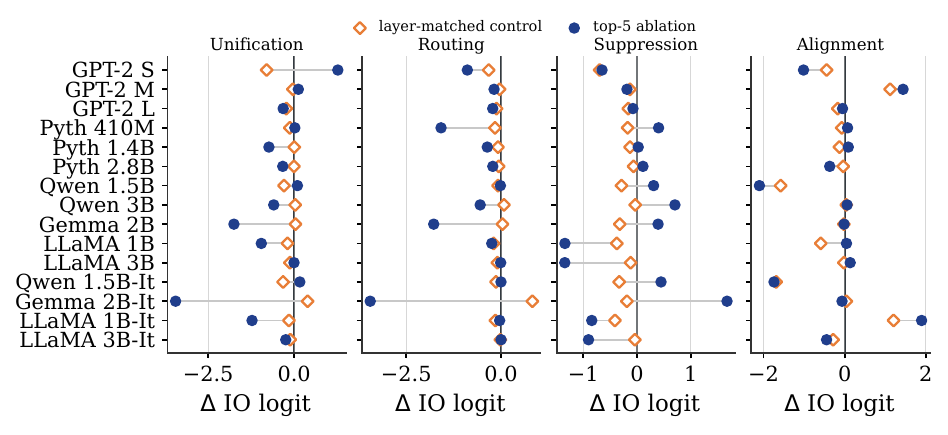}
  \caption{Causal ablation of operator-ranked heads on held-out IOI (mean over 5 seeds). Filled dots: change in the correct-token logit when zero-ablating the top five heads per operator ranking (routing ranked by the embedding-space copy score, Equation~\ref{eq:copy-score}); open diamonds: layer-matched random-head control. Dots left of zero indicate removed support for the correct answer.}
  \label{fig:causal}
\end{figure}

The analyses above measure association, not causation. We test if operator-ranked heads cause the predicted behaviour with the ablation design of Section~\ref{sec:eval-setup} (Figure~\ref{fig:causal}).

Ablating the top unification heads reduces the correct-token logit in 10 models, exceeding the layer-matched control in 9 and the magnitude-matched control in 11. For routing, the copy-score ranking selects a head set largely distinct from unification (mean top-5 overlap $0.19$), and ablating it reduces the correct-token logit in 14 models, exceeding both controls in 10. Magnitude and direction vary by family: Gemma-2B-It shows an IO-logit change of $-3.45$ for routing ablation (control: $+0.83$), while GPT-2~Small's unification ablation \emph{raises} the logit ($+1.27$; control: $-0.80$), the clearest counter-example to the prediction, matching the backup behaviour documented for this circuit, where removing supportive heads recruits compensatory ones \citep{wang2022interpretability}. Its routing ablation behaves as predicted ($-0.89$).

For suppression, the result is weaker: the ablation exceeds the layer-matched control in only 5 models, and the predicted release of the wrong token appears in 7, with the margin moving oppositely or negligibly in the remaining 8. The weight score thus does not reliably isolate task-specific suppression (see Limitations): it measures a head's total subtractive capacity, while which content that capacity acts on depends on the residual state. Task-agnostic structure does not guarantee task-relevant suppression. Re-ranking suppression and unification heads by the targeted DLA margin on the discovery split restores causal reliability on the held-out split: the margin moves in the predicted direction in all models for both rankings (mean $+2.07$ suppression, $-1.86$ unification), and DLA-ranked unification ablation reduces the correct-token logit in every model (mean $-1.08$), at the cost of requiring task labels. \emph{Weight-based rankings suffice for routing and unification, while identifying task-specific suppression requires activation evidence.}

\subsection{Suppression--Alignment Coupling (P2)}\label{sec:eval-coupling}
CPC treats suppression as a response to conflicts exposed through alignment: ablating upstream alignment heads should reduce the summed suppressive contribution of downstream suppressor heads (design in Section~\ref{sec:eval-setup}).

\begin{table}[t]
\centering
\scriptsize
\setlength{\tabcolsep}{3.5pt}
\begin{tabular}{@{}lrrrl@{}}
\toprule
\textbf{Model} & \textbf{$\Delta$ contra} & \textbf{$\Delta$ rand.} & \textbf{$\Delta$ no-conf.} & \textbf{P2} \\
\midrule
GPT-2 Small & -1.39$^{*}$ & -0.43 & -1.74 & \textsc{part.} \\
GPT-2 Medium & +0.02 & -0.01 & -0.02 & \textsc{agai.} \\
GPT-2 Large & -0.01$^{*}$ & -0.00 & -0.02 & \textsc{part.} \\
Pythia-410M & -0.21$^{*}$ & -0.01 & -0.33 & \textsc{part.} \\
Pythia-1.4B & -0.10$^{*}$ & -0.05 & -0.14 & \textsc{part.} \\
Pythia-2.8B & -0.34$^{*}$ & +0.01 & -0.44 & \textsc{part.} \\
Qwen-1.5B & -0.06 & -0.08 & -0.07 & \textsc{agai.} \\
Qwen-3B & +0.02 & -0.12 & +0.46 & \textsc{agai.} \\
Gemma-2B & -0.11$^{*}$ & -0.01 & -0.11 & \textsc{part.} \\
LLaMA-1B & +0.06 & -0.11 & +0.02 & \textsc{agai.} \\
LLaMA-3B & -0.29$^{*}$ & +0.09 & -0.28 & \textsc{supp.} \\
Qwen-1.5B-It & -0.09 & -0.13 & -0.13 & \textsc{agai.} \\
Gemma-2B-It & -0.02$^{*}$ & -0.01 & -0.02 & \textsc{part.} \\
LLaMA-1B-It & -0.15$^{*}$ & -0.08 & -0.23 & \textsc{part.} \\
LLaMA-3B-It & -0.22$^{*}$ & -0.00 & -0.23 & \textsc{part.} \\
\bottomrule
\end{tabular}
\caption{Suppression-alignment coupling (mean over 5 seeds): change in summed suppressive contribution when ablating alignment heads on contradiction prompts ($^{*}$: exceeds the layer-matched random-head control, Holm-corrected $p{<}0.05$), the random-head control, and the same intervention on matched no-conflict prompts. P2: pre-specified verdict (\textsc{supp.}/\textsc{part.}/\textsc{agai.}).}
\label{tab:exp2a}
\end{table}

Table~\ref{tab:exp2a} reports the per-model results. Ablating alignment heads reduces the summed suppressive contribution on contradiction prompts in 12 models, and \emph{the reduction exceeds the layer-matched random-head control with Holm-corrected significance in 10} (95\% bootstrap CIs exclude zero). The conflict-specificity control is rarely passed: only LLaMA-3.2-3B shows a reduction that is significantly weaker on matched no-conflict prompts, and in the remaining significant models the same intervention reduces suppressive activity comparably with and without conflict. Under the pre-specified rule, the outcome is 1 \textsc{support}, 9 \textsc{partial}, and 5 \textsc{against}. Alignment ablation thus propagates to downstream suppressor heads in most architectures, aligned with compositional coupling through the residual stream, but the propagation reflects generic upstream dependence, not conflict-specific signalling. The CPC prediction survives only in its weaker, non-selective form, with suppressor heads appearing to rely on a broad set of upstream writes instead of a dedicated conflict channel. Conflict-specific signalling, if present, is not separable by zero-ablation at this granularity.

\subsection{Coherence Under Contradiction (P3)}\label{sec:eval-contradiction}
The framework predicts that an explicit contradiction depresses the layerwise coherence proxy $\hat{\Cfunc}_{\mathrm{link}}^{(\ell)}$ (Equation~\ref{eq:coherence-proxy}), with the largest disruption near the layers where the model's alignment and suppression activity concentrates (Section~\ref{sec:eval-layered}).

\begin{figure}[t]
  \centering
  \includegraphics[width=\linewidth]{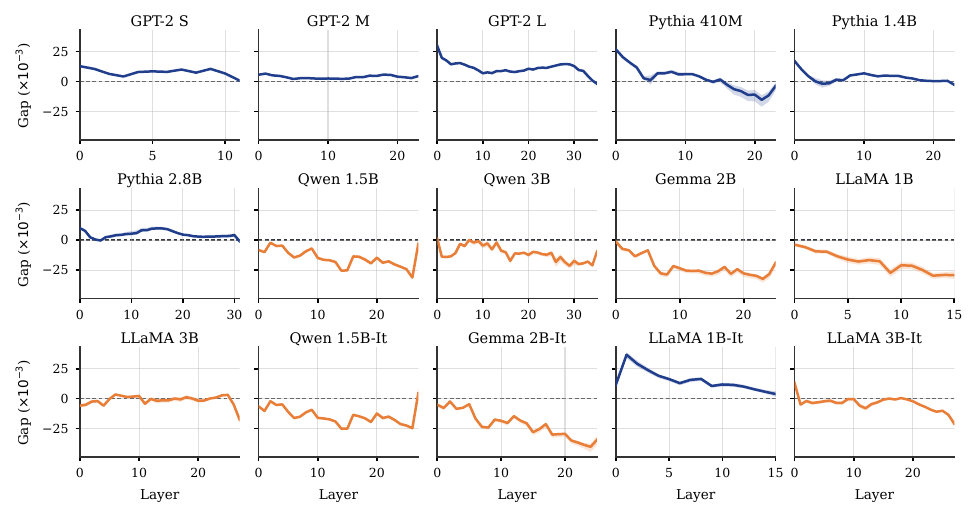}
  \caption{Layerwise coherence gap (consistent minus contradictory) for all 15 models. Lines give the mean over 5 seeds, bands the range across seeds, and colour the sign of the mean gap (navy positive, orange negative). GPT-2 and Pythia show the predicted positive gap, Gemma, Qwen, and three of the four LLaMA variants show negative gaps, and LLaMA-1B-Instruct follows the positive families.}
  \label{fig:exp3-coherence}
\end{figure}

Figure~\ref{fig:exp3-coherence} shows the layerwise coherence gap for all 15 models. \emph{The gap is significant (paired $t$-test, Holm-corrected) in 14 models, but its sign is architecture-dependent}: GPT-2 and Pythia show the predicted positive gap (magnitudes ${\sim}10^{-3}$ to $10^{-2}$), whereas Gemma, Qwen, and three of the four LLaMA variants show a comparable \emph{negative} gap: contradictory prompts \emph{raise} pairwise similarity. The exception is LLaMA-3.2-3B, whose gaps are small and do not survive correction. LLaMA-1B-Instruct, the fourth variant, shows a positive gap and follows the GPT-2 and Pythia pattern. The pre-specified co-location rule gives 10 \textsc{support}, 4 \textsc{mixed} (both Gemma variants, GPT-2 Small, and Qwen-1.5B-Instruct), and 1 \textsc{against}. The directional prediction is assessed separately below. Co-location is exact and seed-stable in all Pythia models: the gap and the suppression profile peak in the same third at every size and seed.

\paragraph{Controls and robustness}
Three analyses qualify the interpretation (designs in Section~\ref{sec:eval-setup}). First, lexically matched non-contradictory pairs produce gaps one to two orders of magnitude smaller in every model: the effect is driven by the contradiction substitution. Second, the shuffled position-pair control reproduces the raw gap almost exactly in 14 models: the raw proxy is a global state marker of contradiction, not a measure of within-prompt relational structure. Third, on the whitened proxy the directional prediction holds: the gap is positive in all models and significant in 14 (Holm-corrected, exception LLaMA-1B-Instruct). This attributes the raw sign flip to residual-stream covariance. One possible explanation is that contradiction engages a shared response component in these families, raising all pairwise similarities at once and masking the relational effect; removing the shared covariance recovers the predicted decrease in every family. As with the signature scores, a global measure mainly reflects the family's overall geometry, and the predicted relational effect appears only once that shared structure is removed. The logit-lens measure agrees: contradiction worsens the rank of the coherent continuation in all models, with the largest changes concentrated in the late third. On the naturalistic Wikipedia set the gap is positive in 10, a weak out-of-template replication. The ascent conjecture (Equation~\ref{eq:ascent}) behaves consistently: on mixed prompts the proxy's correlation with depth is non-negative in 14 models (Spearman $\rho$ up to $0.93$ in Qwen and LLaMA), although it is close to zero for GPT-2~Large and Pythia-410M and negative for GPT-2~Medium.

\begin{table}[t]
\centering
\footnotesize
\setlength{\tabcolsep}{3pt}
\begin{tabular}{@{}lcccccc@{}}
\toprule
\textbf{Pair} & \textbf{Late s+r} & \textbf{Early a} & \textbf{Ratio} & \textbf{Ind.\ $r$} & \textbf{Pres.} & \textbf{Verdict} \\
\midrule
Gemma-2B & 0.006 & 0.017 & 0.33 & 0.982 & yes & \textsc{partial} \\
LLaMA-1B & 0.060 & 0.051 & 1.18 & 0.992 & yes & \textsc{support} \\
LLaMA-3B & 0.055 & 0.040 & 1.36 & 0.991 & yes & \textsc{support} \\
Qwen-1.5B & 0.002 & 0.003 & 0.61 & 0.998 & yes & \textsc{partial} \\
\bottomrule
\end{tabular}
\caption{Base vs.\ instruction-tuned reweighting (mean over 5 seeds): late-half suppression-and-routing signature shift, early-half alignment shift, their ratio, the induction-head correlation between variants, and the outcome of the one-sided preservation test (threshold $r{=}0.9$). Verdicts follow the pre-specified P4 rule.}
\label{tab:exp4}
\end{table}

\subsection{Base vs.\ Instruction-Tuned Models (P4)}\label{sec:eval-reweight}
CPC treats post-training as a possible reweighting of operator structure already present in the base model (Section~\ref{sec:rlhf}). Table~\ref{tab:exp4} summarises the four pairs. Induction-head matching passes the pre-specified preservation test in all four pairs ($r{=}0.98$--$1.00$; one-sided test against $r{=}0.9$, $p{<}0.05$): \emph{induction-head score structure is strongly preserved after post-training, while a generic late concentration of the signature shift is not observed}. Under the pre-specified operator-specific rule (Section~\ref{sec:eval-setup}), the late suppression-and-routing shift exceeds the early alignment shift in both LLaMA pairs but not in Gemma or Qwen, giving two \textsc{support} and two \textsc{partial} verdicts. The operator-specific reweighting prediction thus varies by family, while induction-head score preservation is consistent across the four pairs. The preservation result matches Section~\ref{sec:rlhf}, where the KL constraint keeps post-training close to the pretrained computation: post-training adjusts behaviour while leaving the base circuits supporting in-context prediction intact. Where the adjustment concentrates appears to follow each family's depth organisation, echoing the architecture dependence seen throughout.

\subsection{Cross-Task Operator Stability}\label{sec:eval-crosstask}
If CPC operators are functional kinds, the same heads should play similar roles on different tasks (design in Section~\ref{sec:eval-setup}). Suppression is consistently more task-stable than unification: mean cross-task suppression $r{=}0.68$ versus unification $r{=}0.45$, with suppression exceeding $0.5$ in 12 models and Gemma-2B the most stable (mean $r{=}0.99$). Among the models evaluated on IOI and factual recall, unification stability is higher (mean $r{=}0.58$). The three-task average underestimates cross-task consistency where greater-than has low accuracy. The prediction holds fully in 6 models (both operators $r{>}0.5$), partially in 6, and fails on both operators in 3. The pattern suggests \emph{suppression is a consistent, task-general operation, while unification is more task-dependent}. CPC currently treats both roles as equally task-general; the evidence supports this only for suppression.

\paragraph{Overall pattern}
Across architectures, the most stable findings concern the functional operator axes, causal effects of routing and unification, cross-task suppression, and induction-score preservation. Architecture dependence is stronger in signature profile count, operator timing, raw contradiction geometry, and the location of post-training changes. Role-level regularities are consequently more stable across families than their depth and geometric expression. \emph{Role-level claims can be stated architecture-independently, while depth-level claims should be validated per family.}

\section{Related Work}\label{sec:related-work}
\paragraph{Mechanistic interpretability}
Circuit analysis provides CPC's empirical basis, including QK/OV decomposition \citep{elhage2021framework}, induction heads \citep{olsson2022context,chen2024induction}, the IOI circuit \citep{wang2022interpretability}, copy suppression \citep{mcdougall2023copysuppression}, feedforward memories \citep{geva2021transformer}, and analyses of greater-than, factual recall, and component reuse \citep{hanna2023greaterthan,geva-etal-2023-dissecting,meng2022rome,merullo2024reuse}. Sparse autoencoders, feature circuits, and attribution graphs link interpretable features to causal subgraphs \citep{cunningham2023sparse,templeton2024scaling,marks2024sparse,lindsey2025biology}. Path patching, automated circuit discovery, and edge-attribution methods locate the causally relevant components on which such findings rest \citep{goldowskydill2023pathpatching,conmy2023acdc,hanna2024have}, and the component types they identify are consistent with the operator roles CPC defines. Closest to CPC's aim, modular-circuit approaches seek a global vocabulary of reusable task-agnostic subgraphs \citep{he2025modcirc}; CPC instead defines its vocabulary through graded functional roles tied to a coherence objective, under which heads may mix roles instead of partitioning into discrete modules. Circuit analyses of logical reasoning find dedicated structure for propositional inference and syllogisms, including a middle-term suppression circuit \citep{hong2024implies,kim2025reasoning}; CPC treats such contradiction handling as one instance of its suppression role. The operators could be formalised as causal-abstraction variables, with interchange interventions or causal scrubbing verifying proposed head-to-operator assignments \citep{geiger2025causal,chan2022causal}, or induced from circuit evidence through inductive-logic theory construction \citep{aljaafari2026circuit}.

\paragraph{Theoretical frameworks for in-context learning and post-training}
Bayesian and algorithmic frameworks treat transformers as implicit inference machines or estimators \citep{xie2021explanation,akyurek2023icl_bayes,von2022transformers,garg2022what_can_tfs_learn}; CPC differs by naming the component-level functions that these frameworks abstract away: alignment, unification, suppression, and routing. Preference-training objectives change model behaviour \citep{ouyang2022instruct,bai2022constitutional,rafailov2023dpo}, and mechanistic studies suggest that fine-tuning can enhance or reweight existing circuits \citep{prakash2024finetuning,jain2024mechanistically,ruscio2026pretraining}, with instruction-conditioned suppression \citep{wang2023forbiddenfacts} providing a circuit-level example for CPC's reweighting interpretation. \citet{boleda-2025-llms} combines symbolic-like and continuous computation, compatible with parts of CPC but without a central role for contradiction suppression.

\section{Conclusion}\label{sec:conclusion}
We proposed Coherentist Probabilistic Compositionalism (CPC), an interpretive framework that describes transformer computation through four operator roles involved in coherence construction and readout, and evaluated its predictions on 15 models from five architecture families. The vocabulary identifies axes of head variation that hold across architectures, while the number of separable signature profiles, the depth at which suppression peaks, and the way contradictions register all depend on the family. CPC provides a shared vocabulary for transformer mechanisms and post-training effects, with predictions that should be stated conditionally on architecture. Verifying the operator assignments through causal abstraction and encoding the task-generality asymmetry between suppression and unification are natural next steps.

\section*{Limitations}\label{sec:limitations}
CPC is an interpretive framework, not a mechanistic proof: the four operators are not claimed to be exhaustive, the treatment covers decoder-only English models, and the predicted failure regimes (ambiguity overload, depth overload, preference distortion) remain untested. The main measurement caveat concerns the coherence proxy, which conflates semantic coherence with residual-stream geometry: the raw gap's sign reverses between families and is reproduced by cross-prompt position pairs, so it is a global state marker of contradiction, not a measure of within-prompt relational structure (Section~\ref{sec:eval-contradiction}). Whitening and the logit-lens trajectory provide convergent directional support, but a proxy that isolates relational coherence pair-specifically remains future work, and the ascent conjecture (Equation~\ref{eq:ascent}) is weak or absent in the GPT-2 family and Pythia-410M.

The suppression evidence is the weakest. The weight score measures total subtractive OV mode weight, a property of the matrix, and does not reliably isolate task-specific suppression (layer-matched control exceeded in 5/15 models); task-targeted DLA rankings restore causal reliability (Section~\ref{sec:eval-causal}) but require task labels. The evaluation rests on synthetic tasks with a single naturalistic negation set (gap positive in 10/15 models), greater-than covers only the six GPT-2 and Pythia models for tokeniser reasons, and the base-vs-tuned comparison covers four pairs. The coupling test (P2) uses the validated IOI circuit only for GPT-2~Small, with automatically detected heads elsewhere, and its conflict-specificity control passed only in LLaMA-3.2-3B, so the P2 verdicts rest on the random-head control; other operationalisations of ``the suppressed alternative'' remain untested.

\section*{Ethics Statement}\label{sec:ethics}
CPC is a mechanistic interpretive framework. Its predictions concern internal representations, not system deployment. We do not anticipate direct misuse risks from this analytical vocabulary; however, improved understanding of suppression and instruction-following circuits may inform both the design and circumvention of safety mechanisms.

\appendix

\appendixsection{The CPC Functional and Coherentism}\label{app:coherence}
The CPC functional in Equation~\ref{eq:coherence-functional} adapts the constraint-based view of Section~\ref{sec:coherence-csp} to transformer computation. Grounded fragments $(m,g)$ play the role of elements in the constraint formulation, while the relational structure $G$ specifies how these fragments can be connected. The local score $\Scoh(g\mid m,G)$ rewards a grounding that fits the developing interpretation, corresponding to positive compatibility between elements. The kernel $\kappa$ penalises incompatible grounded fragments, corresponding to negative constraints. The additional coverage term rewards interpretations that incorporate more of the available input, preventing coherence from being increased simply by retaining a small set of mutually compatible fragments.

Three differences separate CPC from the discrete constraint-satisfaction formulation. First, support is graded: fragments receive continuous compatibility scores instead of belonging only to accepted or rejected sets. Second, the problem is incremental. New linguistic material is introduced token by token, so the interpretation must be updated as the context grows. Third, the optimisation is implicit. CPC does not claim that a transformer explicitly represents Equation~\ref{eq:coherence-functional} or solves a constraint-satisfaction problem. The stronger claim is instead formulated as the coherence-ascent conjecture of Section~\ref{sec:cpc-ascent}: computation across depth can be interpreted as repeated refinement of the current representational state towards configurations with greater coherence (see Figure~\ref{fig:configuration}).

The four CPC operators describe functional contributions to this refinement. \emph{Alignment} identifies candidate relations between fragments. \emph{Unification} incorporates mutually compatible information into the developing representation. \emph{Suppression} reduces support for incompatible alternatives, paralleling inhibitory competition in the constraint-satisfaction formulation. \emph{Routing} then carries the resulting information to positions where it can affect the readout. The correspondence is functional, not an identification of linguistic coherence theories with transformer mechanisms.

This connection also clarifies the status of the empirical coherence proxy in Equation~\ref{eq:coherence-proxy}. Similarity between internal representations measures one possible consequence of compatible information becoming more closely represented, and is related to semantic-similarity approaches to discourse coherence. It does not measure the complete functional in Equation~\ref{eq:coherence-functional}. In particular, it does not directly identify discourse relations, entity continuity, or incompatibility between specific groundings. For this reason, CPC treats the proxy as an empirical diagnostic of representational coherence, while the coherence state and functional provide the broader theoretical construct.

The distinction between measurement and construct in this account parallels the levels of description in physical theory. Circuit-level measurements are analogous to instrument readings, which sit close to the observable. The operator labels and the functional are analogous to constructs such as energy and entropy, which organise many observations at a higher level of description and are not read from any single measurement.

\appendixsection{Formal Definitions}\label{app:defs}
A fragment $m{=}(U, F, C)$ specifies a position set $U \subseteq \{1, \ldots, T\}$, a content label $F$ from a finite vocabulary of semantic and syntactic types, and compositional constraints $C$. A grounding is a partial function $g : U \rightharpoonup V$ into the relational graph $G{=}(V, E)$, admissible when it satisfies $C$. Two grounded fragments are compatible when their assignments agree on all shared positions and jointly satisfy both constraint sets. The contradiction kernel $\kappa \geq 0$ is symmetric with $\kappa{=}0$ on compatible pairs. The local score $\Scoh(g \mid m, G) \in (0,1]$ is a content-compatibility factor times the geometric mean of pairwise relational compatibilities $\psi\bigl(g(u), g(u')\bigr)$; for nodes with activation vectors, $\psi = \epsilon_\psi + (1-\epsilon_\psi)\bigl(1+\cos(r_v,r_{v'})\bigr)/2$, bounded in $(0,1]$.

\appendixsection{Task Definitions}\label{app:tasks}

The three tasks of Section~\ref{sec:eval-setup} share a next-token format: the context is a token sequence $c=\{\mathrm{tok}_1, \mathrm{tok}_2, \dots, \mathrm{tok}_n\}$, the model predicts its continuation, and an example enters the experiments only when the prediction is correct under the task's criterion. Multi-token targets are scored on their first token.

\noindent\paragraph{Indirect Object Identification (IOI)}
This task requires the model to predict the indirect object of a transfer clause, given a context that introduces two names and repeats one of them as the subject, as in \codeword{When Mary and John went to the store, John gave a drink to}. Formally, the context $c$ introduces names $A$ and $B$ and repeats $B$ as the subject of the final clause, and the model seeks to produce the non-repeated name $A$ such that
\begin{equation}
    A = \arg\max_{t \in \mathcal{V}} P(t \mid c),
\end{equation}
where $\mathcal{V}$ denotes the model's vocabulary. Predictions are deemed correct only if the predicted token matches the first token of $A$. The repeated name $B$ provides the wrong-token contrast for the logit margin used in the ablation experiments (Section~\ref{sec:eval-setup}). Prompts instantiate 15 template frames over a pool of roughly 100 names, with places and objects varied.

\noindent\paragraph{Greater-Than Comparison}
This task requires the model to complete a year interval consistently, given a context that states a start year and the century prefix of the end year, as in \codeword{The war lasted from the year 1732 to the year 17}. With $\mathrm{YY}$ the two-digit suffix of the start year, the model seeks a two-digit continuation $\hat{y}$ such that
\begin{equation}
    \hat{y} = \arg\max_{t \in \mathcal{V}} P(t \mid c), \qquad \mathrm{YY} < \operatorname{val}(\hat{y}) \leq 99,
\end{equation}
where $\operatorname{val}(\cdot)$ reads a two-digit token as a number. Predictions are deemed correct only if the predicted token is a single-token two-digit year satisfying the inequality; this requirement excludes nine models from the task (Section~\ref{sec:eval-setup}). Examples are drawn from an exhaustive noun-by-year grid over the years 1102--1898, excluding suffixes 00 and 99.

\noindent\paragraph{Factual Recall}
This task requires the model to predict the object of a subject-relation pair, given a natural-language prompt that expresses the subject and the relation, as in \codeword{The Eiffel Tower is located in the city of}. Formally, the model seeks to produce the gold object $o$ such that
\begin{equation}
    o = \arg\max_{t \in \mathcal{V}} P(t \mid c).
\end{equation}
Prompts and gold objects come from the CounterFact set \citep{meng2022rome}, and predictions are deemed correct only if the predicted token matches the first token of $o$.

\appendixsection{Parameter and Threshold Choices}\label{app:params}

The constants below were fixed in the experiment plan before the confirmatory runs, and none was tuned against held-out outcomes. Choices fall into three classes: values preregistered with an explicit rationale, values matched to validated reference circuits, and conventional defaults. Every stochastic component (clustering, null draws, bootstraps, control-head samples) runs at seeds $\{0,\ldots,4\}$, with means and ranges reported across seeds.

\noindent\paragraph{Prompt counts and splits}
The target of 500 model-correct examples per task balances stable per-head averages and paired tests against the cost of running 15 models with 5 seeds under interventions; candidate prompts are oversampled fourfold before correctness filtering. The 50/50 discovery/held-out split gives selection and verdict computation equal power, and hashing at generation time fixes the assignment so no example migrates between splits. The contradiction pool of 3{,}854 pairs exhausts the unique fills of the generation templates; 500 pairs per run matches the per-task target.

\noindent\paragraph{Weight-space analysis}
The cluster range $k \in \{2,\ldots,7\}$ brackets the hypothesised four-role organisation from both sides, with $k{=}2$ the minimal nontrivial split, letting the silhouette peak fall below, at, or above four instead of being forced towards it. Fifty $k$-means restarts guard against local minima, and assignments are stable across the five seeds. The two silhouette nulls use 200 draws each, giving a $p$-value resolution of $1/201$, finer than the reporting level of $p{<}0.01$; the 100 random weight projections give the 95\% baseline quantile a resolution of one draw.

\noindent\paragraph{Activation measures}
DLA is computed on the model's top-10 predicted tokens so that attribution is restricted to the model's own candidate set, where logit contributions bear on the prediction; widening the set dilutes the measure over tokens the model never considers. All layerwise quantities are computed at every layer; the layer third is the unit of summary and testing, not of measurement (per-layer profiles appear in Figure~\ref{fig:exp3-coherence}). Testing at thirds serves three purposes. The CPC predictions are phase-level claims (early alignment, mid-network unification and suppression, late routing); the test resolution matches the claim. Thirds are comparable across architectures whose depths range from 12 to 36 layers, where individual layer indices do not align. Three tests per model also keep the Holm family small, whereas per-layer testing would inflate it by an order of magnitude and reduce power. The whitening shrinkage of $\alpha{=}0.1$ addresses the rank-deficient per-batch covariance (positions are far fewer than dimensions); the value is a fixed conventional intensity, and the eigenvalue floor guards numerical stability.

\noindent\paragraph{Head-set sizes}
The detection sizes for P2 mirror the validated GPT-2~Small IOI sets: five alignment heads (three duplicate-token plus two previous-token) and six suppressors (four S-inhibition plus two negative name-movers), so that automatic detection selects sets of the same size as the reference circuit. The causal ablations remove the top five heads per operator, applied uniformly: the set is comparable to identified circuit sizes and remains a small fraction of the head budget of even the smallest model (five of GPT-2~Small's 144 heads). The ablation-effect correlation uses the union of the top-10 heads per operator because a correlation over all heads would be dominated by the large bulk of heads with near-zero scores and near-zero effects.

\noindent\paragraph{Statistical thresholds}
Holm correction at $\alpha{=}0.05$ within each experiment family follows convention. Permutation tests and bootstrap intervals use 10{,}000 draws, placing the Monte Carlo standard error near $0.002$ at $p \approx 0.05$. The preservation margin $r{=}0.9$ was preregistered and uses a pre-specified preservation threshold of $r=0.9$: shared variance must exceed $0.81$, so preservation is never inferred from a non-significant difference. The cross-task stability threshold $r{>}0.5$ is a conventional boundary for a moderate-to-strong correlation, fixed before analysis.

\appendixsection{Contradiction Prompt Templates}\label{app:contradiction-templates}
Contradiction prompts are drawn from a pool of 3{,}854 globally unique consistent/contradictory pairs, generated once (seed 42). Each pair shares identical surface structure up to a key substitution that introduces or removes a contradiction. Pool 1 (3{,}333 pairs) instantiates four named-agent IOI-style frame pairs over a 100-name pool with place and object slots. Pool 2 (221 pairs) covers seven contradiction types (factual, temporal, causal, logical, narrative, spatial, quantitative) via parameterised generators. Pool 3 (300 pairs) crosses 20 agent-competence frames with 15 role pairs, without name placeholders. A representative pool-1 pair contrasts ``When \emph{A} and \emph{B} went to the store, \emph{A} stayed in the store, and \emph{A} gave a drink to'' with the same frame containing ``\emph{A} immediately left the store''; the other pools follow the same substitution pattern. A separate naturalistic set pairs Wikipedia lead sentences with pronoun-coreferent continuations, negating the first sentence's copula in the contradictory variant; it provides the out-of-template replication of Section~\ref{sec:eval-contradiction}.

\begin{acknowledgments}
This is an example of acknowlwdgment. This is a sample paper presented jost for the coding of different elements of a document in \LaTeX\ using this package file. We thank all readers.
\end{acknowledgments}


\bibliographystyle{compling}
\bibliography{COLI_template}

\end{document}